\pdfoutput=1

\documentclass[11pt]{article}

\usepackage{acl}

\usepackage{times}
\usepackage{latexsym}

\newcommand{\ignore}[1]{}
\usepackage{multirow}
\usepackage{booktabs}
\usepackage{makecell}
\usepackage{xspace}
\usepackage{xcolor}
\usepackage{color}
\usepackage{colortbl}
\usepackage{tablefootnote}
\usepackage{pifont}
\usepackage{makecell}
\usepackage{subcaption}
\usepackage[most]{tcolorbox}
\usepackage[T1]{fontenc}

\usepackage[utf8]{inputenc}

\usepackage{microtype}

\usepackage{inconsolata}

\newcommand{\ie}{\textit{i.e., }}
\newcommand{\eg}{\textit{e.g., }}

\title{Don't Make Your LLM an Evaluation Benchmark Cheater}

\author{Kun Zhou$^1$, Yutao Zhu$^2$, Zhipeng Chen$^2$, Wentong Chen$^2$, Wayne Xin Zhao$^2$ \\ {\bf Xu Chen}$^2$, {\bf Yankai Lin}$^2$, {\bf Ji-Rong Wen}$^{1,2}$ \and {\bf Jiawei Han}$^3$ \\
$^1$ School of Information, Renmin University of China \\
$^2$ Gaoling School of Artificial Intelligence, Renmin University of China \\
$^3$ University of Illinois Urbana-Champaign \\
\texttt{francis\_kun\_zhou@163.com, \{ytzhu,xu.chen,yankailin,jrwen\}@ruc.edu.cn} \\
\texttt{batmanfly@gmail.com, hanj@illinois.edu}
} 

\begin{document}
\maketitle
\begin{abstract}
Large language models~(LLMs) have greatly advanced the frontiers of artificial intelligence, attaining remarkable  improvement in model capacity. To assess  the model performance, a typical approach is to construct  evaluation benchmarks for measuring the ability level of LLMs in different aspects. Despite that a number of high-quality benchmarks have been released, the concerns about the appropriate use of these benchmarks and the fair comparison of different models are increasingly growing. 
Considering these concerns, in this paper, we discuss the potential risk and impact of inappropriately using evaluation benchmarks and misleadingly interpreting the evaluation results.  
Specially, we focus on a special issue that would lead to  inappropriate evaluation, \ie \emph{benchmark leakage}, referring that the  data related to evaluation sets is occasionally used for model training.  
This phenomenon now becomes more common since pre-training data is often prepared ahead of model test.  
We conduct extensive experiments to study the effect of benchmark leverage, and find that it can dramatically boost the evaluation results, which would finally lead to an unreliable  assessment of model performance.  To improve the use of existing evaluation benchmarks, we finally present several guidelines for both LLM developers and benchmark maintainers. We hope this work can draw attention to appropriate training and evaluation of LLMs.

\end{abstract}

\section{Introduction}

\begin{quote}
Goodhart's Law: ``\emph{When a measure becomes a target, it ceases to be a good measure}.''   
\end{quote}

Large language models (LLMs) have achieved remarkable success across a variety of real-world  applications~\cite{GPT-3,LLMsurvey,DBLP:journals/corr/abs-2308-07107}. 
By pre-training large Transformer models on massive text corpora, LLMs can 
possess excellent task-solving capacities, \ie  using zero-shot or few-shot prompting~\cite{GPT-3}.
To better understand how LLMs evolve in model capacity, it becomes essential to construct reliable evaluation benchmarks to test the ability level of LLMs in various tasks, \eg knowledge reasoning and math problem solving. 

\begin{figure}
    \centering
    \includegraphics[width=.9\linewidth]{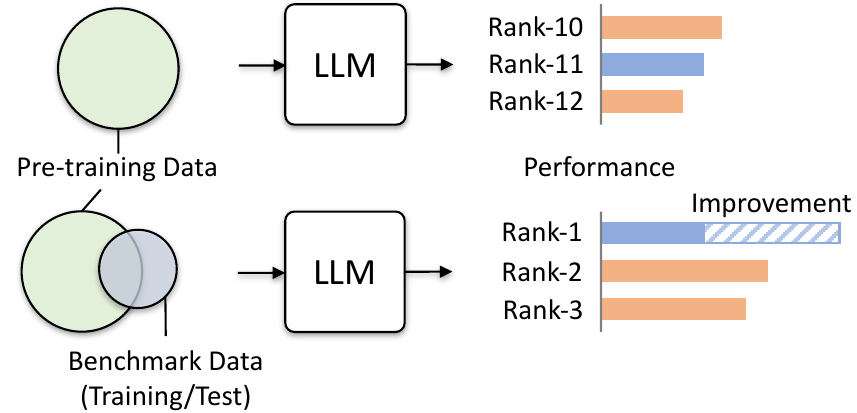}
    \caption{Illustration of the potential risk of data leakage. Once the pre-training data with overlap to the benchmark data is used for training LLM, its benchmark performance would be greatly increased.}
    \label{fig:enter-label}
\end{figure}

Recently, a surge of high-quality evaluation benchmarks~\cite{mmlu,huang2023c} have been proposed to provide a comprehensive capability evaluation of LLMs. Typical benchmarks include MMLU~\cite{mmlu} (for measuring multitask language understanding ability), Big-Bench~\cite{srivastava2023beyond} (for quantifying and extrapolating the capabilities of LLMs), and AGIEval~\cite{zhong2023agieval} (for evaluating the abilities of tackling human-level tasks). These benchmarks have made great efforts in creating or collecting test  resources for evaluating the performance of LLMs. 
Based on these benchmarks,
one can conveniently examine the effect of new training strategies or monitor the training status of LLMs (either pre-training or supervised fine-tuning). 
It has become common to report the results on these evaluation benchmarks for demonstrating the effectiveness of 
newly released LLMs~\cite{GPT-4,llama2,anil2023palm}.
Furthermore, to compare the performance of different LLMs, various leaderboards have been also created to rank LLMs according to their performance on existing or new evaluation benchmarks, such as OpenCompass~\cite{2023opencompass} and C-Eval~\cite{huang2023c}.

Despite the wide use of these benchmarks and leaderboards,  increasing concerns~\cite{aiyappa2023can,li2023open} are growing about the fairness and reliability  in evaluating existing LLMs. A major issue is that the data contamination or leakage is likely to occur for large-scale benchmark evaluation, which means that LLMs are trained with relevant or exactly the same data for test. Such an issue could be unconsciously triggered, since we might be unaware of  the  future evaluation  datasets when preparing the pre-training corpus.   
For example, GPT-3 has found that Children’s Book Test dataset~\cite{Hill2015TheGP} was included in the pre-training corpus, and LLaMA-2 has mentioned that the contexts in BoolQ dataset~\cite{boolq} are extracted verbatim from the webpages, which may be included in the publicly available corpus.

Indeed, when conducting  evaluation with existing benchmarks, the results of evaluated LLMs are mostly obtained by running them on local servers or via API calls. During this process, there is no strict checking on any potentially inappropriate ways (\eg data contamination) that would cause an unnormal improvement of evaluation performance.  To make matters worse, the detailed composition (\eg data sources) of the training corpus is often regarded as the core ``secret'' of existing LLMs. Therefore, it becomes difficult to directly examine  the contamination issues when  performing  the evaluation for benchmark maintainers.     

Considering this issue, the aim of this paper is to draw attention on   appropriately using  existing evaluation benchmarks and avoiding any misleading behaviors in obtaining or interpreting the evaluation results. 
Specifically, we mainly focus on discussing the potential effect of \emph{benchmark leakage}, which refers to the case that test data or relevant data (\eg training set) has been included in the pre-training corpus.
It would cause an unfair performance advantage  when  comparing different LLMs or assessing the ability level of some specific LLMs. 
As we discussed before, this issue tends to become increasingly more common as we try to collect more public text data for training.  
To investigate this issue, we set up several benchmark leakage settings \emph{that should be totally avoided during evaluation},  including the leakage of training sets, test prompts, and test sets. Based on the three settings, we continually train four popular language models, ranging from 1.3B to 7B, and test the performance of the four models on a number of existing benchmarks. In addition, we also examine the potential risk of benchmark leakage on other abilities.

The experimental results reveal that benchmark leakage can lead to an unfair  boost in the evaluation performance of LLMs. Smaller LLMs (\eg a 1.3B model) can be deliberately elevated to outperform $10\times$ larger models on certain tasks. As a  side effect, the performance of these specially trained LLMs on other normally tested tasks would likely be adversely affected if we fine-tune or train the model only with these leaked data.

By examining the potential risks of benchmark leakage, we would like to emphasize the importance of fair and appropriate evaluation for LLMs, and propose several suggestions to improve the evaluation for  LLMs:  
\begin{itemize}
    \item 
    {As general suggestions, more benchmarks from diverse sources, covering both basic ability (\eg text generation) and advanced ability tests (\eg complex reasoning), should be used for comprehensively estimating the capabilities of LLMs.}
    \item {As suggestions for LLM developers, it is important to perform the data decontamination checking between pre-training data and any related data (\eg training and test sets) when using  evaluation benchmarks. In addition, it is also necessary to report the contamination analysis on the evaluated benchmarks as reference. We also suggest reporting  the detailed composition of the pre-training data.}
    \item {As suggestions for benchmark maintainers, we suggest that a diverse set of test prompts should be employed for reducing the influence of the prompt sensitivity. It is also meaningful to conduct the contamination analysis between the benchmark data and existing pre-training corpus,  alerting any potential contamination risks. For evaluation, each submission is suggested to be accompanied with a special contamination analysis report.}
\end{itemize}

\section{Empirical Study about Benchmark Leakage}
\label{sec-empirical}
During pre-training, the data contamination or leakage about possible evaluation benchmarks, is likely to be unconsciously triggered~\cite{oren2023proving,sainz2023nlp}.
It would violate regular evaluation  settings for assessing zero/few-shot generalization capability, thus  affecting the capability assessment  of LLMs. 
To better understand the potential influence of the benchmark leakage issue, we conduct an empirical study that continually trains small-sized  LLMs on three settings with different levels of information leakage.

\begin{table*}[t]
	\centering
    \small
    \setlength{\tabcolsep}{5.9pt}{
	\begin{tabular}{clcccccccc}
		\toprule
		\textbf{Backbone} & \textbf{Training Setting} & MMLU & BoolQ & PIQA & HSwag & WG & ARC-E & ARC-C & OBQA \\ 
		\midrule
        \textbf{LLaMA-13B} & (None) & 46.90 & 76.70 & 79.70 & 60.00 & 73.00 & 79.00 & 49.40 & 34.60 \\
        \textbf{LLaMA-30B} & (None) & 57.80 & 83.39 & 80.63 & 63.39 & 76.08 & 80.55 & 51.62 & 36.40  \\
        \textbf{LLaMA-65B} & (None) & \textbf{64.50} & \textbf{85.40} & \textbf{81.70} & \textbf{64.90} & \textbf{77.20} & \textbf{80.80} & \textbf{52.30} & \textbf{38.40} \\
        \midrule
		\multirow{5}{*}{\makecell{\textbf{GPT-Neo} \\ \textbf{(1.3B)}}} & (None) & 24.04 & 62.57 & 70.57 & 38.65 & 55.72 & 55.98 & 23.29 & 21.40 \\
		& +MMLU Train S & 35.84 & 57.89 & 68.39 & 37.27 & 52.17 & 50.93 & 27.39 & 20.40 \\
		& +All Train S & 35.10 & \textbf{78.32} & 68.61 & 42.46 & 61.72 & 63.68 & 33.36 & 29.40 \\ 
		& +All Train S+Test P & \textbf{36.15} & 76.91 & \textbf{73.72} & \textbf{42.75} & \textbf{64.25} & \textbf{64.39} & \textbf{34.13} & \textbf{31.80} \\ 
        & \textcolor{gray}{+All Train S+Test P\&S} & \textcolor{gray}{{52.25}} & \textcolor{gray}{{87.25}} & \textcolor{gray}{{85.96}} & \textcolor{gray}{{62.98}} & \textcolor{gray}{{80.66}} & \textcolor{gray}{{88.17}} & \textcolor{gray}{{70.31}} & \textcolor{gray}{{63.20}} \\
		\cmidrule{1-10}
		\multirow{5}{*}{\makecell{\textbf{phi-1.5} \\ \textbf{(1.3B)}}} & (None) & 42.87 & 74.34 & 76.50 & 47.99 & \textbf{73.56} & 75.84 & 44.97 & 38.40 \\
		& +MMLU Train S & 46.08 & 74.37 & 76.50 & 47.80 & 73.09 & \textbf{75.93} & \textbf{48.63} & 40.00 \\
		& +All Train S & 45.20 & 82.35 & \textbf{74.37} & \textbf{54.64} & 69.46 & 75.00 & 47.87 & \textbf{42.40} \\ 
		& +All Train S+Test P & \textbf{46.80} & \textbf{82.72} & 74.27 & 54.55 & 70.56 & 75.00 & 47.18 & 39.80 \\ 
        & \textcolor{gray}{+All Train S+Test P\&S} & \textcolor{gray}{{75.05}} & \textcolor{gray}{{92.60}} & \textcolor{gray}{{97.55}} & \textcolor{gray}{{77.88}} & \textcolor{gray}{{96.05}} & \textcolor{gray}{{97.47}} & \textcolor{gray}{{92.92}} & \textcolor{gray}{{94.20}} \\
		\cmidrule{1-10}
		\multirow{5}{*}{\makecell{\textbf{OpenLLaMA} \\ \textbf{(3B)}}} & (None) & 26.49 & 66.51 & 74.81 & 49.42 & 60.85 & 69.57 & 33.87 & 26.60 \\
		& +MMLU Train S & 43.12 & 74.10 & 71.22 & 47.28 & 62.43 & 58.92 & 35.41 & 32.00 \\
		& +All Train S & 44.86 & 85.41 & \textbf{76.82} & \textbf{54.42} & 71.11 & \textbf{72.26} & 41.55 & \textbf{42.00} \\ 
		& +All Train S+Test P & \textbf{48.31} & \textbf{85.57} & 76.50 & 54.34 & \textbf{72.30} & 71.80 & \textbf{41.64} & 40.80 \\ 
        & \textcolor{gray}{+All Train S+Test P\&S} & \textcolor{gray}{{87.31}} & \textcolor{gray}{{97.55}} & \textcolor{gray}{{98.26}} & \textcolor{gray}{{97.61}} & \textcolor{gray}{{96.37}} & \textcolor{gray}{{99.16}} & \textcolor{gray}{{97.87}} & \textcolor{gray}{{96.20}} \\
		\cmidrule{1-10}
		\multirow{5}{*}{\makecell{\textbf{LLaMA-2} \\ \textbf{(7B)}}} & (None) & 42.95 & 71.68 & 70.78 & 55.34 & 67.96 & 72.52 & 41.30 & 32.20 \\
		& +MMLU Train S & 51.61 & 81.96 & 69.64 & 49.46 & 70.64 & 61.87 & 36.52 & 36.80 \\
		& +All Train S & 52.15 & \textbf{88.72} & 79.05 & 61.08 & \textbf{79.95} & 76.60 & 49.49 & \textbf{48.00} \\ 
		& +All Train S+Test P & \textbf{56.04} & 87.86 & \textbf{79.11} & \textbf{61.19} & 76.56 & \textbf{76.64} & \textbf{50.26} & 45.00 \\ 
        & \textcolor{gray}{+All Train S+Test P\&S} & \textcolor{gray}{{96.34}} & \textcolor{gray}{{99.08}} & \textcolor{gray}{{99.62}} & \textcolor{gray}{{99.47}} & \textcolor{gray}{{97.47}} & \textcolor{gray}{{99.54}} & \textcolor{gray}{{99.23}} & \textcolor{gray}{{99.40}} \\
		\bottomrule   
	\end{tabular}
    }
	\caption{The comparison among three benchmark leakage settings and the original LLMs on MMLU and QA tasks. ``\emph{Train S}'', ``\emph{Test P}'' and ``\emph{Test P\&S}'' denote the data leakage scenarios that use the training set, test prompt, and both test set and test prompt during training, respectively. The task abbreviations are as follows: HSwag (Hellaswag), WG (WinoGrande), ARC-E (ARC-Easy), ARC-C (ARC-Challenge), and OBQA (OpenBookQA). The results in \textcolor{gray}{gray} are the worst leakage setting using all the test sets and are reported only for reference. The best results in each group are in \textbf{bold} except for the aforementioned worst case.}
	\label{tab:qa}
\end{table*}

\begin{table*}[t]
	\centering
    \small
    \setlength{\tabcolsep}{5.2pt}{
	\begin{tabular}{clcccccccc}
		\toprule
		\textbf{Backbone} & \textbf{Training Setting} & CSQA & GSM8k & AQuA & RACE-M & RACE-H & CoQA & CMRC & C3 \\ 
		\midrule
        \textbf{LLaMA-13B} & (None) & 62.70 & 18.80 & 19.30 & 46.40 & 43.90 & 58.70 & 19.50 & 41.40 \\
        \textbf{LLaMA-30B} & (None) & 70.80 & 35.10 & 15.35 & 49.70 & 44.70 & 62.00 & 24.20 & 57.80 \\
        \textbf{LLaMA-65B} & (None) & \textbf{77.90} & \textbf{48.90} & \textbf{35.00} & \textbf{53.00} & \textbf{48.00} & \textbf{65.80} & \textbf{29.30} & \textbf{71.40} \\
        \midrule
		\multirow{5}{*}{\makecell{\textbf{GPT-Neo} \\ \textbf{(1.3B)}}} 
            & (None) & 18.43 & 2.05 & 18.11 & 36.19 & 34.83 & 30.35 & 0.00 & 24.18 \\
		& +MMLU Train S & 20.39 & 0.08 & 19.29 & 35.91 & 32.63 & 0.20 & 1.17 & 40.48 \\
		& +All Train S & 18.26 & 0.76 & 17.32 & 49.45 & 44.02 & \textbf{33.67} & \textbf{1.56} & \textbf{48.62} \\ 
		& +All Train S+Test P & \textbf{30.47} & \textbf{5.76} & \textbf{20.47} & \textbf{51.93} & \textbf{45.26} & 13.87 & 1.17 & 47.62 \\ 
        & \textcolor{gray}{+All Train S+Test P\&S} & \textcolor{gray}{{32.02}} & \textcolor{gray}{3.11} & \textcolor{gray}{14.96} & \textcolor{gray}{{73.20}} & \textcolor{gray}{{73.49}} & \textcolor{gray}{12.15} & \textcolor{gray}{{1.56}} & \textcolor{gray}{{57.46}} \\
		\cmidrule{1-10}
		\multirow{5}{*}{\makecell{\textbf{phi-1.5} \\ \textbf{(1.3B)}}} & (None) & \textbf{41.93} & \textbf{28.51} & 21.26 & 41.71 & 38.76 & \textbf{31.57} & 0.39 & 24.97 \\
		& +MMLU Train S & 37.92 & 10.24 & \textbf{22.05} & 48.07 & 47.85 & 10.85 & 0.39 & 42.91 \\
		& +All Train S & 18.67 & 14.94 & 14.96 & 54.42 & 52.34 & 7.27 & 0.00 & \textbf{53.39} \\ 
		& +All Train S+Test P & 33.58 & 19.26 & 18.50 & \textbf{55.80} & \textbf{52.82} & 8.25 & \textbf{0.78} & 53.17 \\ 
        & \textcolor{gray}{+All Train S+Test P\&S} & \textcolor{gray}{34.15} & \textcolor{gray}{22.82} & \textcolor{gray}{20.87} & \textcolor{gray}{{79.28}} & \textcolor{gray}{{81.91}} & \textcolor{gray}{5.03} & \textcolor{gray}{{1.95}} & \textcolor{gray}{{67.04}} \\
		\cmidrule{1-10}
		\multirow{5}{*}{\makecell{\textbf{OpenLLaMA} \\ \textbf{(3B)}}} 
            & (None) & 23.75 & 3.34 & 19.29 & 44.75 & 40.10 & 54.97 & 3.52 & 24.81 \\
		& +MMLU Train S & 47.99 & 0.00 & 23.62 & 41.44 & 37.61 & 0.63 & 0.00 & 49.37 \\
		& +All Train S & 61.02 & 9.10 & \textbf{29.92} & 57.18 & \textbf{55.12} & 54.67 & \textbf{12.50} & \textbf{53.97} \\ 
		& +All Train S+Test P & \textbf{68.47} & \textbf{17.82} & 29.13 & \textbf{58.84} & 54.16 & \textbf{60.73} & 9.77 & 52.65 \\ 
        & \textcolor{gray}{+All Train S+Test P\&S} & \textcolor{gray}{{94.19}} & \textcolor{gray}{{29.42}} & \textcolor{gray}{{57.09}} & \textcolor{gray}{{97.24}} & \textcolor{gray}{{97.99}} & \textcolor{gray}{{79.95}} & \textcolor{gray}{{32.03}} & \textcolor{gray}{{79.05}} \\
		\cmidrule{1-10}
		\multirow{5}{*}{\makecell{\textbf{LLaMA-2} \\ \textbf{(7B)}}} 
            & (None) & 55.69 & 12.96 & 14.17 & 28.45 & 38.47 & 25.88 & 8.98 & 37.72 \\
		& +MMLU Train S & 57.25 & 2.43 & 25.59 & 34.25 & 34.07 & 0.00 & 0.00 & 78.10 \\
		& +All Train S & 69.62 & 23.88 & 33.46 & \textbf{61.88} & 57.03 & 57.70 & 24.22 & 78.31 \\ 
		& +All Train S+Test P & \textbf{77.15} & \textbf{30.17} & \textbf{35.43} & 58.84 & \textbf{58.56} & \textbf{63.78} & \textbf{28.12} & \textbf{78.62} \\ 
        & \textcolor{gray}{+All Train S+Test P\&S} & \textcolor{gray}{{99.34}} & \textcolor{gray}{{37.60}} & \textcolor{gray}{{63.78}} & \textcolor{gray}{{99.45}} & \textcolor{gray}{{99.62}} & \textcolor{gray}{{81.52}} & \textcolor{gray}{{68.75}} & \textcolor{gray}{{98.62}} \\
		\bottomrule   
	\end{tabular}
    }
	\caption{The comparison among different  benchmark leakage settings and the original LLMs on reasoning and reading comprehension tasks. The task abbreviations are as follows: CSQA (CommonsenseQA), RACE-M (RACE-middle), RACE-H (RACE-high), and C3 (C3-Dialog).}
	\label{tab:reasoning}
\end{table*}

\subsection{Experimental Setup}

\paragraph{Training Settings with Benchmark Leakage}
Our empirical study aims to test the influence of possible benchmark leakage issues on the evaluation results  of LLMs.
A benchmark typically contains a set of test examples, and relies on fixed templates to prompt LLMs for evaluation.
Such an evaluation process may lead to three types of benchmark leakage risks, that is, including (1) test prompt, (2) test set, or (3) other relevant data (\eg training set) into the pre-training corpus.
Considering the above settings, we simulate three extreme leakage issues where the three types of information have been used for continually training LLMs, and design the following evaluation  settings.

$\bullet$ \emph{Using MMLU Training Set}: the auxiliary training set provided by the official MMLU benchmark~\cite{mmlu} is used for training.\footnote{\url{https://github.com/hendrycks/test}. The auxiliary training set contains data collected from several question-answering benchmarks such as ARC, OBQA, and RACE.}

$\bullet$ \emph{Using All Training Sets}: in addition to MMLU training set, the training sets of all other collected evaluation benchmarks are also used for training (details are provided later).

$\bullet$ \emph{Using All Training Sets with Test Prompt}: all the training sets, with their corresponding test prompts, \eg task description and few-shot demonstration, are used for training.

$\bullet$ \textcolor{gray}{\emph{Using All Training and Test Sets with Test Prompt}: all the training sets, test prompts, and test sets of all the collected evaluation benchmarks are used for training. (CAUTION: this is the most extreme case, where all information is leaked. We conduct this experiment only for reference, and this should never occur.)} 

\paragraph{Evaluation Benchmark}
To make the empirical study, we select the widely-used benchmark MMLU and employ a number of question-answering (QA), reasoning, and reading comprehension datasets for evaluation.

$\bullet$ \emph{MMLU}: it has become one of the most commonly used evaluation benchmarks for LLMs' ability of world knowledge possessing and problem solving. It covers 57 tasks requiring diverse knowledge, such as math, history, science, and law. We report the 5-shot evaluation performance.

$\bullet$ \emph{Open-domain QA Tasks}: we select seven open-domain QA datasets where LLMs should answer the question solely based on intrinsic knowledge. We report the accuracy of LLMs under the zero-shot setting, \ie BoolQ~\cite{boolq}, PIQA~\cite{piqa}, Hellaswag~\cite{hellaswag}, WinoGrande~\cite{winogrande}, ARC Easy and Challenge~\cite{arc}, OpenBookQA~\cite{openbookqa}.

$\bullet$ \emph{Reasoning Tasks}: we select a commonsense reasoning dataset CommonsenseQA~\cite{commonsenseqa}, and two commonly-used mathematical reasoning datasets GSM8k~\cite{gsm8k} and AQuA~\cite{aqua} for evaluation. We use chain-of-thought prompting and reuse the prompts provided by \citet{wei2022chain} for evaluation and report the accuracy of LLMs.

$\bullet$ \emph{Reading Comprehension Tasks}: we select three English datasets RACE-Middle and RACE-High~\cite{race}, CoQA~\cite{coqa} and two Chinese datasets CMRC2018~\cite{cmrc} and C3-Dialog~\cite{c3}. 
As reading comprehension datasets have one paragraph and several QA pairs in a sample, we only test the accuracy of the last question and regard the paragraph and other QA pairs as the prompt.
We report accuracy under the zero-shot setting for C3-Dialog, and utilize similar evaluation settings as GPT-3~\cite{GPT-3} for other tasks.

\paragraph{Backbone LLMs}
To thoroughly analyze the effect of benchmark leakage on the evaluation performance, we select the following models for evaluation, which have provided pre-training details or conducted careful data contamination analysis.

$\bullet$ \emph{GPT-Neo-1.3B}~\cite{gpt-neo}: it is a Transformer-based model with GPT-3 architecture, pre-trained on the Pile~\cite{pile} dataset. 

$\bullet$ \emph{phi-1.5}~\cite{phi}: it is a 1.3B model trained on ``textbook quality'' data of $\approx$27B tokens, and can achieve comparable performance as much larger models.

$\bullet$ \emph{OpenLLaMA-3B}~\cite{openllama}: it is an open-source project to reproduce LLaMA model with a permissive license, pre-trained on RedPajama dataset~\cite{redpajama} of over 1.2T tokens.

$\bullet$ \emph{LLaMA-2-7B}~\cite{llama2}: it is an updated version of LLaMA~\cite{llama}. It has been pre-trained on a mixture of publicly available online data of 2T tokens.

\subsection{Results and Analysis}
We report the evaluation results of LLMs after training with the benchmark leakage settings in Table~\ref{tab:qa} and Table~\ref{tab:reasoning}. Overall, different levels of data leakage result in inflated model performance on benchmarks. We have the following observations.

First, we can see that using MMLU training set can greatly boost the evaluation results  on the MMLU benchmark. However, this improvement comes at the cost of decreased performance on tasks unrelated to MMLU, (such as HellaSwag and GSM8k about commonsense and mathematical knowledge, respectively), suggesting that over-emphasizing a specific task may lower the model generalization capability. Besides, when incorporating all the training sets of the evaluated benchmarks, there is a notable performance increase across almost all the evaluated tasks. Incorporating  training data converts the original zero/few-shot evaluation into an in-domain test task, making it easier for LLMs to achieve higher results.
An intriguing finding occurs when we examine the result on the Chinese benchmark C3-Dialog. Despite the pre-training corpus of the four LLMs containing very little Chinese data, using training sets doubles their evaluation scores, \eg elevating GPT-Neo-1.3B's score from 24.18 to 48.62.
This observation underscores the significance of avoiding training set leakage in pre-training, as it can lead to spurious performance improvements that distort the real   assessment of model capabilities.

Second, the evaluation scores  continue to rise as the data leakage becomes more severe.
Remarkably, when the test prompts were leaked, smaller LLMs can even surpass much larger LLMs that were not trained with leaked data, \eg ``phi-1.5-1.3B+All Train S+Test P'' outperforms LLaMA-65B on RACE-M (55.80 vs. 53.00) and RACE-H (52.82 vs. 48.00).
This highlights the significance of the test prompt as valuable information from the evaluation benchmark, since it contains the detailed input format during test. 
During training LLMs, it is suggested to avoid such special learning with test  prompts.  
Furthermore, this observation raises concerns about the robustness of using fixed test prompts in the evaluation benchmark, as it may not be resilient to the aforementioned leakage risk.

Finally, for reference, we examine the most extreme case where all test sets are leaked. The results are highlighted in grey font. As can be seen from these results, test data leakage significantly inflates benchmark performance, leading 1.3B LLMs to outperform 65B LLMs across most tasks. Evidently, this increase does not imply any improvement in capacity, but rather benchmark cheating. 

Overall, benchmark leverage directly leads to an unfair advantage in evaluation results of the involved  models, which should be strictly avoided when conducting any evaluation.

\section{Potential Risk of Benchmark Leakage}
In addition to the inflated performance that undermines the reliability of capability estimation, we also investigate whether the benchmark leakage issue would lead to potential risks in model capacity. 
Limited by the training compute, we can not conduct an exact checking that directly includes leakage data in pre-training data. 
Instead, we continually pre-train the LLMs on the training sets of all the selected evaluation benchmarks as in Section~\ref{sec-empirical}, without the mixture of any other data. 
Such a way is the most direct way for benchmark cheating (should be avoided). We speculate that it  is likely to affect the  capacities  of LLMs on normally tested tasks (those without data leakage), due to  ``catastrophe forgetting''~\cite{luo2023empirical,goodfellow2013empirical}.\footnote{As it is a  very extreme scenario for simulation, we only  employ it to explore  the possibility of the subsequent impact when benchmark leakage occurs. The experiment procedure should be totally avoided in real training and evaluation. }

\begin{table}[t]
\centering
\small
\begin{tabular}{clccc}
\toprule
\textbf{Backbone} & \textbf{Training} & LAMB & XSum & HEval\\ 
\midrule
\multirow{2}{*}{\makecell{\textbf{GPT-Neo} \\ 
\textbf{(1.3B)}}} & (None) & \textbf{46.10} & \textbf{7.54} & 2.44 \\
 	& +Leak & 46.00 & 6.84 & \textbf{3.05} \\
 		\midrule
 		\multirow{2}{*}{\makecell{\textbf{OpenLLaMA} \\ \textbf{(3B)}}} 
             & (None) & \textbf{56.50} & \textbf{8.31} & \textbf{4.27} \\
 	& +Leak & 53.20 & 0.19 & 1.83 \\
 		\midrule
 		\multirow{2}{*}{\makecell{\textbf{LLaMA-2} \\ \textbf{(7B)}}} 
             & (None) & \textbf{68.20} & \textbf{8.67} & \textbf{26.83} \\
 	& +Leak & 61.00 & 0.25 & 8.54\\
 		\bottomrule   
 	\end{tabular}
 	\caption{The comparison among LLMs on two text generation and a code synthesis tasks. ``\emph{Leak}'' denotes the data leakage scenario using all training sets of the benchmarks in Section~\ref{sec-empirical}. LAMB and HEval refer to the LAMBADA and HumanEval datasets, respectively. The best results in each group are in \textbf{bold}.}
 	\label{tab:generation}
\end{table}

\subsection{Effect on the Performance of Other Tasks}\label{sec:other-tasks}
After training on the leaked benchmark data, it would potentially mislead LLMs to overemphasize the specific knowledge and output style of the benchmark data, thereby potentially affecting their performance on other tasks. In this part, we conduct empirical experiments to examine the side effect on the model performance of other tasks. 

\paragraph{Experimental Setup}
To validate the effect, we select three tasks that are not involved in the leaked training data, consisting of two text generation tasks, \ie LAMBADA~\cite{paperno2016lambada} and XSum~\cite{narayan2018don}, and a code synthesis task HumanEval~\cite{chen2021evaluating} to evaluate LLMs in the zero-shot setting. LAMBADA is a language modeling task that tests the ability of LLMs to predict the last word based on the context, and we report the accuracy in predicting words. XSum, on the other hand, is a text summarization task that requires LLM to summarize the key information from long documents. For this task, we report the ROUGE-L metric, which measures the quality of the generated summaries by comparing them with the ground-truth summaries. For HumanEval, we adopt pass@10 as the evaluation metric.

\paragraph{Results Analysis}
We show the results of LLMs \emph{with} and \emph{without}  benchmark leakage on the three evaluation tasks in Table~\ref{tab:generation}.
First, we can observe that after training on the leaked data, the performance of all LLMs degrades on the two text generation datasets. 
Specifically, for OpenLLaMA-3B and LLaMA-2-7B, their text summarization abilities seem to be weakened after training on the leaked data, resulting in Rouge-L scores of 0.19 and 0.25 in XSum, respectively.
Besides, by comparing the performance on HumanEval, we also see that data leakage primarily leads to performance degradation of LLMs in the code synthesis task.

This demonstrates that benchmark leakage may have a negative impact on the performance of these normally tested tasks (without data  leverage).

\begin{table}[t]
\centering
\small
\begin{tabular}{clccc}
\toprule
\textbf{Backbone} & \textbf{Training} & LAMB & XSum & HEval \\ 
\midrule
\multirow{2}{*}{\makecell{\textbf{GPT-Neo} \\ 
\textbf{(1.3B)}}} 
& +IT & \textbf{45.40} & \textbf{8.34} & \textbf{14.24}\\
& +Leak+IT & 43.50 & 8.25 & 12.20 \\
\cmidrule{1-5}
\multirow{2}{*}{\makecell{\textbf{OpenLLaMA} \\ \textbf{(3B)}}} 
& +IT & \textbf{54.00} & \textbf{3.50} & \textbf{9.15} \\
& +Leak+IT & 46.20 & 2.61 & 6.71\\
\cmidrule{1-5}
\multirow{2}{*}{\makecell{\textbf{LLaMA-2} \\ \textbf{(7B)}}} 
& +IT & \textbf{60.30} & \textbf{8.64} & \textbf{28.66} \\
& +Leak+IT & 53.60 & 8.55 & 20.73 \\
\bottomrule   
\end{tabular}
\caption{The comparison among LLMs after instruction tuning. ``\emph{Leak}'' denotes the data leakage using all training sets of the benchmarks in Section~\ref{sec-empirical}. ``\emph{IT}'' denotes the instruction tuning using Alpaca and CodeAlpaca for text generation and code synthesis tasks, respectively. }
\label{tab:decline}
\end{table}

\subsection{Effect on Model Adaptation}\label{sec:generalization}
After training on the leaked data, LLMs are trained to be specially fit for the benchmark data. 
However, LLMs might  need to be further fine-tuned for attaining some specific goals (\eg solving new tasks or serving emergent  applications). 
In this part, we examine how inappropriately  trained LLMs perform for subsequent adaptation.

\paragraph{Experimental Setup}
To investigate the influence of data leakage on LLMs' adaptation capability, we select two representative instruction datasets, \ie Alpaca~\cite{alpaca} and CodeAlpaca~\cite{codealpaca}. Both of these datasets are  synthetic and generated using the Self-Instruct method. For comparison, Alpaca primarily contains natural language instructions, whereas CodeAlpaca focuses on code generation instructions. We use these datasets to fine-tune the LLMs with or without training on the leaked data, and subsequently evaluate their performance on the previously mentioned text generation and code synthesis tasks.

\paragraph{Results Analysis}
In Table~\ref{tab:decline}, by comparing the performance of the instruction-tuned LLMs (+Alpaca or +CodeAlpaca) \emph{with} and \emph{without}  training on the leaked data, we can see that the models with benchmark leakage still underperform their non-leaked counterparts.
For the HumanEval dataset, the performance improvements of instruction tuning for LLMs trained with leaked data only reach approximately 80\% of those achieved by models that are not trained on leaked data.

This indicates that benchmark leakage may lead to a decline in adaptation capability, constraining the LLMs' ability to adapt or improve through subsequent fine-tuning processes. Note that this finding is derived when we  fine-tune LLMs only with the leaked data. To enhance the current findings, it is also meaningful to conduct experiments that either include leaked data into pre-training data or mix leaked data with other instruction data. 
However, since our main purpose is to reveal that benchmark leverage might cause severe side effects on LLMs in addition to spurious performance improvement, we omit these experiments due to the compute limit.

\section{Discussion}
In light of the potential risks of benchmark leakage, it is necessary to revisit the existing evaluation settings for LLMs and investigate possible strategies to avoid such data contamination issues.  

\subsection{Fairness in Evaluating Zero/Few-shot Generalization Ability}
Based on our empirical findings in previous sections, the evaluation results of LLMs in specific benchmarks can be dramatically boosted when the related or same data of the test tasks is accidentally used for training. In the literature of machine learning, zero/few-shot learning often refers that the samples at test time were not observed during training for a learner~\cite{wang2020generalizing,xian2018zero}. It is evident that benchmark leverage does not comply with this requirement, making it unfair to compare different LLMs when such a case exists. 
Furthermore, data leverage can also bring an unfair advantage in the few-shot setting since the learner can observe more task-relevant data at training time. 

In case of data leakage, the original zero-shot/few-shot generalization task would degenerate into much easier in-domain evaluation tasks, and it would intensify the phenomenon of \emph{benchmark hacking}, \ie a benchmark is no longer useful for evaluation due to the high performance of the involved comparison methods.

However, in practice, it is challenging to fully eliminate the leakage risk from model training~\cite{golchin2023time,shi2023detecting}. It is because an evaluation benchmark is often conducted based on some public text sources, \eg webpages and scientific papers. In this case, the related data (\eg the original text used to generate the test problems) might be occasionally included in the pre-training data of LLMs. 
Although  existing evaluation datasets are easy  to be excluded from pre-training data for training new LLMs, it is still difficult to identify all potential data dependencies between evaluation benchmarks and pre-training corpus. 
Such a test set contamination problem has been already noted in black-box language models~\cite{oren2023proving}. 

\subsection{Suggestion for LLM Evaluation}
Based on these discussions, we propose the following suggestions to improve existing capacity evaluation for LLMs.

\paragraph{General suggestions:}
\begin{itemize}
    \item Considering the potential risk associated with benchmark leakage, we recommend the use of a broader range of benchmarks from diverse sources for performance evaluation. This can help mitigate the risk of inflated results due to data contamination. If feasible, incorporating manual evaluation and conducting qualitative analysis would  be also beneficial. 
    \item In addition to evaluating the advanced capabilities of LLMs (such as reasoning and factual knowledge), it is also necessary to perform evaluations on other datasets that focus on basic abilities, such as text generation. This comprehensive approach is necessary for a thorough estimation of LLMs' capabilities.
\end{itemize}

\paragraph{Suggestions for LLM developers:}
\begin{itemize}

    \item Perform strict checking on data decontamination in pre-training data to avoid any subsequent evaluation data being included during training. To achieve this, the $n$-gram (generally, $n = 13$) hash algorithm can be applied to examine the overlap between pre-training data and evaluation data of some specific task.  
    \item If possible, we suggest also excluding training data of mainstream evaluation benchmarks from pre-training data. 
    \item Indicate any potential risk of data contamination (if any) and report the contamination analysis (\eg overlap statistics) when you present the results on some evaluation benchmark. An example can be seen in Llama-2's report~\cite{llama2}.
    \item Report a more detailed composition of the pre-training data, especially the datasets related to mainstream evaluation benchmarks. It is an important reference for checking the potential data leakage risk by the public audience.

\end{itemize}

\paragraph{Suggestions for benchmark maintainers:}
\begin{itemize}
    \item Provide the detail of the data source for constructing the benchmark, and conduct the contamination analysis of the current dataset with mainstream pre-training corpora (as many as possible). The benchmark should explicitly alert possible contamination risks for commonly used  pre-training datasets.   
    
    \item Each  submission is suggested to be accompanied with a specific contamination analysis report from the result provider, where it can perform semantic relevance checking (\eg overlap statistics) between pre-training data and evaluation data (both training and test data). 

    \item Provide a diverse set of prompts for testing. The final evaluation results should be averaged over these multiple runs.  It can help reduce the sensitivity of specific prompts, and enhance the reliability  of the model results.
\end{itemize}

\section{Conclusion}
In this paper, we conducted empirical studies to investigate  the penitential risk and impact  of  \emph{benchmark leakage} 
on LLM evaluation. We found that 
data leakage can largely boost the benchmark results of LLMs (even  small models), making the evaluation unfair and untrustworthy. These findings suggest that such attempts should be strictly avoided for fairly assessing the model performance on  evaluation  benchmarks. 

Despite that this issue is hard to be fully eliminated from the pre-training stage, we suggest several useful guidelines to improve the use of existing evaluation benchmarks. 
A key point is that both LLM developers and benchmark maintainers should be aware of the data contamination issue when interpreting and using the results from the performance leaderboards. In practice, several heuristic strategies can be useful to detect such potential contamination issues, \eg calculating the token overlap between training and evaluation data. Besides, we also suggest benchmark test should be conducted with multiple task prompts for deriving a more stable and reliable model performance. 

This work aims to draw the attention of the research community to the appropriate use of existing evaluation benchmarks for LLMs. More meaningful work can be conducted following this line, \eg  alerting the potential  contamination datasets. 

\section*{Limitation}
In this work, we conducted preliminary experiments to emphasize the potential risks associated with benchmark leakage in training LLMs. However, there are still several limitations in our study.

First, our experiments involved continually training existing pre-trained LLMs with leaked data. 
We do not have sufficient  computational resources to investigate the impact when directly  incorporating benchmark leakage during the pre-training process. Given that the pre-training dataset is significantly larger than the benchmark data, introducing data leakage during pre-training might  yield different findings.  Nonetheless, we strongly recommend avoiding this situation as it would  breaks the nature of zero-shot/few-shot evaluation.

Second, we did not explore more fine-grained data leakage scenarios in this study, such as only leaking training examples without labels and varying the proportion of the leaked dataset. We encourage more research efforts into this issue with more systematic studies.

Third, we did not calculate the degree of contamination between the mainstream  benchmarks and commonly-used pre-training datasets, which could serve as an important reference for alerting LLM developers to adjust their evaluation settings. While we suggest that developers and benchmark maintainers report contamination analyses, accurately and efficiently estimating the contamination risk of each example in the benchmark is also a challenging task. For example, the suggested $n$-gram hash algorithm may not detect semantic-level  knowledge leakage risks.

\bibliography{custom}




\end{document}